\documentclass[10pt,twocolumn,letterpaper]{article}

\usepackage{cvpr}
\usepackage{times}
\usepackage{epsfig}
\usepackage{graphicx}
\usepackage{amsmath}
\usepackage{amssymb}
\usepackage[table,xcdraw]{xcolor}
\usepackage{multirow}
\usepackage{soul}

\usepackage[pagebackref=true,breaklinks=true,letterpaper=true,colorlinks,bookmarks=false]{hyperref}

\cvprfinalcopy 

\def\cvprPaperID{877} 

\ifcvprfinal\fi
\begin{document}

\title{
Instance-Level Salient Object Segmentation
}

\author{Guanbin Li$^{1,2}$ \quad\quad Yuan Xie$^1$ \quad\quad Liang Lin$^1$ \quad\quad Yizhou Yu$^2$\thanks{Corresponding author (email: yizhouy@acm.org).}\vspace{2mm}\\
	$^1$Sun Yat-sen University \quad\quad\quad $^2$The University of Hong Kong\\
	{\tt\footnotesize liguanbin@mail.sysu.edu.cn}, {\tt\small xiey39@mail2.sysu.edu.cn}, {\tt\small linliang@ieee.org}, {\tt\small yizhouy@acm.org}
	\vspace{-0mm}
}


\maketitle

\begin{abstract}
   Image saliency detection has recently witnessed rapid progress due to deep convolutional neural networks. However, none of the existing methods is able to identify object instances in the detected salient regions. In this paper, we present a salient instance segmentation method that produces a saliency mask with distinct object instance labels for an input image. Our method consists of three steps, estimating saliency map, detecting salient object contours and identifying salient object instances. For the first two steps, we propose a multiscale saliency refinement network, which generates high-quality salient region masks and salient object contours. Once integrated with multiscale combinatorial grouping and a MAP-based subset optimization framework, our method can generate very promising salient object instance segmentation results. To promote further research and evaluation of salient instance segmentation, we also construct a new database of 1000 images and their pixelwise salient instance annotations. Experimental results demonstrate that our proposed method is capable of achieving state-of-the-art performance on all public benchmarks for salient region detection as well as on our new dataset for salient instance segmentation.
\end{abstract}


\section{Introduction}
Salient object detection attempts to locate the most noticeable and eye-attracting object regions in images. It is a fundamental problem in computer vision and has served as a pre-processing step to facilitate a wide range of vision applications including content-aware image editing~\cite{avidan2007seam}, object detection~\cite{navalpakkam2006integrated}, and video summarization~\cite{ma2002user}.

\begin{figure}[t]
\centerline{
   \includegraphics[width=0.48\textwidth]{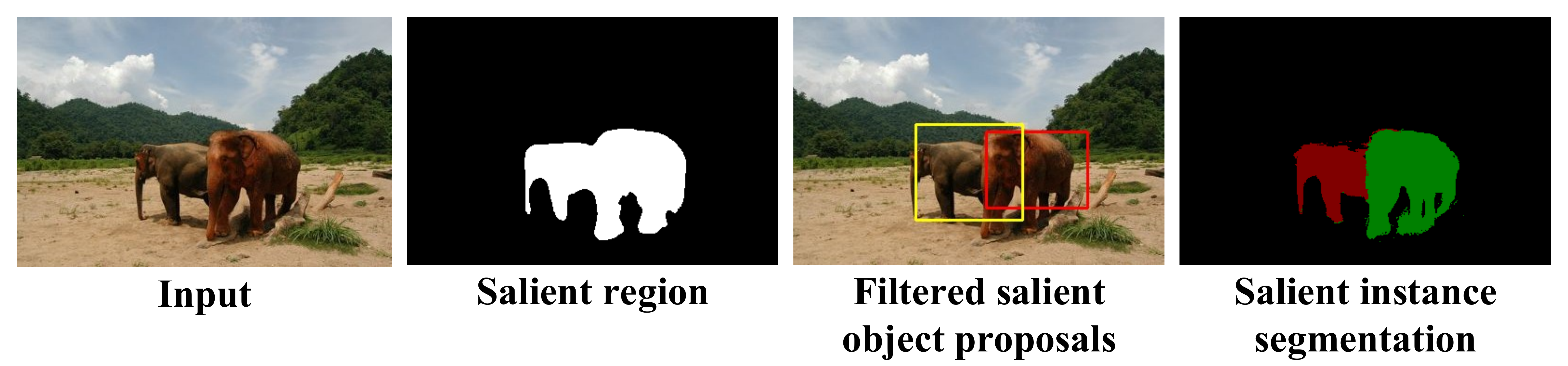}
}
   \caption{An example of instance-level salient object segmentation. Left: input image. Middle left: detected salient region. Middle right: filtered salient object proposals. Right: result of salient instance segmentation. Different colors indicate different object instances in the detected salient region.}
\label{fig:exemplar_salient_instance}
\vspace{-3mm}
\end{figure}

Recently the accuracy of salient object detection has been improved rapidly~\cite{LiYu15,LiYu16,liu2016dhsnet,wang2016saliency} due to the deployment of deep convolutional neural networks. Nevertheless, most of previous methods are only designed to detect pixels that belong to any salient object, i.e. a dense saliency map, but are unaware of individual instances of salient objects. We refer to the task performed by these methods ``salient region detection", as in \cite{zhangunconstrained}. In this paper, we tackle a more challenging task, instance-level salient object segmentation (or {\em salient instance segmentation} for short), which aims to identify individual object instances in the detected salient regions (Fig. \ref{fig:exemplar_salient_instance}). The next generation of salient object detection methods need to perform more detailed parsing within detected salient regions to achieve this goal, which is crucial for practical applications, including image captioning~\cite{karpathy2015deep}, multilabel image recognition~\cite{wei2014cnn} as well as various weakly supervised or unsupervised learning scenarios~\cite{lai2016saliency,chen2015webly}.

We suggest to decompose the salient instance segmentation task into the following three sub-tasks. 1) Estimating binary saliency map. In this sub-task, a pixel-level saliency mask is predicted, indicating salient regions in the input image. 2) Detecting salient object contours. In this sub-task, we perform contour detection for individual salient object instances. Such contour detection is expected to suppress spurious boundaries other than object contours and guide the generation of salient object proposals. 3) Identifying salient object instances. In this sub-task, salient object proposals are generated, and a small subset of salient object proposals are selected to best cover the salient regions. Finally, a CRF based refinement method is applied to improve the spatial coherence of salient object instances.

A number of recent papers have explored the use of fully convolutional neural networks for saliency mask generation~\cite{LiYu16,liu2016dhsnet,wang2016saliency}. Though these methods are efficient and can produce favorable results, they have their own limitations. Most of these methods infer saliency by learning contrast from the internal multi-layer structure of a single VGG network~\cite{wang2016saliency,liu2016dhsnet}. As their output is derived from receptive fields with a uniform size, they may not perform well on images with salient objects at multiple different scales. Though Li {\em et al.}~\cite{LiYu16} combined a multiscale fully convolutional network and a segment-level spatial pooling stream to make up for this deficiency,
the resolution of their final saliency map is only one eighth of the resolution of the original input image, making it infeasible to accurately detect the contours of small salient object instances.

Given the aforementioned sub-tasks of salient instance segmentation, we propose a deep multiscale saliency refinement network, which can generate very accurate results for both salient region detection and object contour detection. Our deep network consists of three parallel streams processing scaled versions of the same input image and a learned attention model to fuse results at different scales from the three streams. The three streams share the same network architecture, a refined VGG network, and its associated parameters. This refined VGG network is designed to integrate the bottom-up and top-down information in the original network. Such information integration is paramount for both salient region detection~\cite{borji2012boosting} and contour detection~\cite{bertasius2015deepedge}.
The attention model in our deep network is jointly trained with the refined VGG network in the three streams.

Given the detected contours of salient object instances, we apply multiscale combinatorial grouping (MCG)~\cite{arbelaez2014multiscale} to generate a number of salient object proposals. Though the generated object proposals are of high quality, they are still noisy and tend to have severe overlap. We further filter out noisy or overlapping proposals and produce a compact set of segmented salient object instances. Finally, a fully connected CRF model is employed to improve spatial coherence and contour localization in the initial salient instance segmentation.

In summary, this paper has the following contributions:
{\flushleft $\bullet$} We develop a fully convolutional multiscale refinement network, called MSRNet, for salient region detection. MSRNet can not only integrate bottom-up and top-down information for saliency inference but also attentionally determine the pixel-level weight of each salient map by looking at different scaled versions of the same image. The proposed network can achieve significantly higher precision in salient region detection than previous methods.

{\flushleft $\bullet$} MSRNet generalizes well to salient object contour detection, making it possible to separate distinct object instances in detected salient regions. When integrated with object proposal generation and screening techniques, our method can generate high-quality segmented salient object instances.


{\flushleft $\bullet$} A new challenging dataset is created for further research and evaluation of salient instance segmentation. 
We have generated benchmark results for salient contour detection as well as salient instance segmentation using MSRNet.

\begin{figure*}[ht]
\begin{center}
   \includegraphics[width=0.91\textwidth]{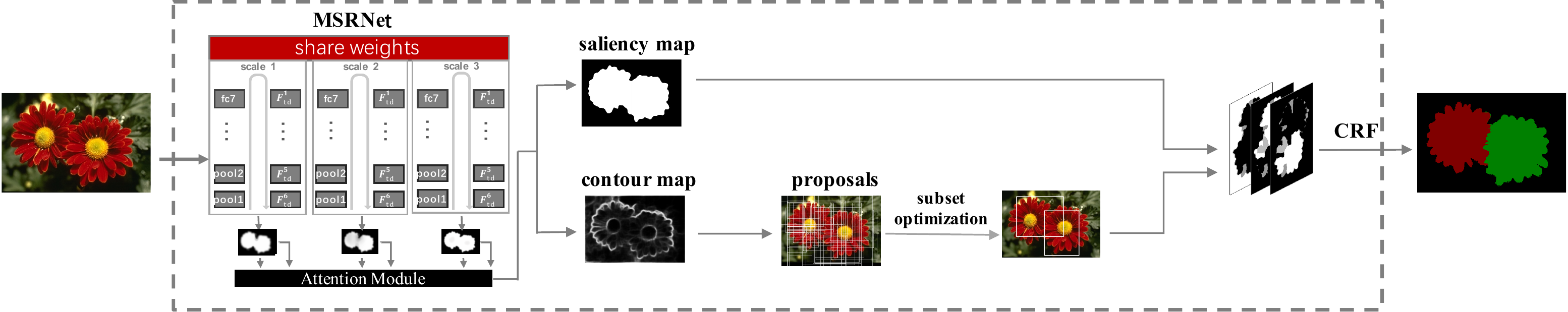}
\end{center}
\vspace{-3mm}
   \caption{Our overall framework for instance-level salient object segmentation.}
\label{fig:architecture}

\end{figure*}

\section{Related Work}
Recently, deep convolutional neural networks have achieved great successes in computer vision topics such as image classification~\cite{krizhevsky2012imagenet,he2015deep}, object detection~\cite{girshick2015fast,redmon2015you} and semantic segmentation~\cite{long2015fully,chen2014semantic}. In this section, we discuss the most relevant work on salient region detection, object proposal generation and instance-aware semantic segmentation.

\subsection{Salient Region Detection}
Traditional saliency detection can be divided into bottom-up methods based on low-level features~\cite{liu2011learning,perazzi2012saliency,cheng2015global} and top-down methods incorporating high-level knowledge~\cite{goferman2012context,li2014secrets,jia2013category}. Recently, deep CNNs have pushed the research on salient region detection into a new phase. Deep CNN based methods can be divided into two categories, segmentation or patch based methods~\cite{LiYu15,wang2016saliency,zhao2015saliency} and end-to-end saliency inference methods~\cite{LiYu16,liu2016dhsnet,wang2016saliency}.
Methods in the former category treat image patches as independent training and testing samples, and are generally inefficient due to redundancy among overlapping patches. To overcome this deficiency, deep end-to-end networks~\cite{LiYu16,liu2016dhsnet,wang2016saliency}
have been developed for saliency inference. Most recently, recurrent neural networks have also been integrated into such networks~\cite{liu2016dhsnet,wang2016saliency}.
Though these end-to-end networks improve both accuracy and efficiency, all of them consider a single scale of the input image and may not perform well on images with object instances at multiple scales. 

\subsection{Object Proposals}
Object proposal generation aims at localizing target objects with a minimum number of object window (or segment) hypotheses. Previous work on this topic can be grouped into two approaches. The first produces a list of object proposal windows, ranked by a measure of objectness (the probability of an image window containing an object)~\cite{zitnick2014edge,cheng2014bing} while the other generates object proposals by merging image segments resulting from multiple levels of segmentation~\cite{arbelaez2014multiscale,van2011segmentation}.
Though they have been widely used as a foregoing step for object detection, they are not tailored for salient object localization. Though Feng {\em et al.}~\cite{feng2011salient} proposed to generate a ranked list of salient object proposals, the overall quality of their result needs much improvement.
Recently, Zhang {\em et al.}~\cite{zhangunconstrained} proposed a MAP-based subset optimization formulation to optimize both the number and locations of detection windows given a set of salient object proposals. However, due to the coarse mechanism they use, their ``filtered'' object windows cannot well match groundtruth objects. In this paper, we generate salient object proposals on the basis of salient object contour detection results.

\subsection{Instance-Aware Semantic Segmentation}
Instance-aware semantic segmentation is defined as a unified task of object detection and semantic segmentation.
This problem was first raised in \cite{hariharan2014simultaneous}, and has been much studied in recent years. It is either formulated as a multi-task learning problem~\cite{hariharan2014simultaneous,dai2015instance} or solved in an end-to-end integrated model~\cite{romera2015recurrent,dai2016instance}. Inspired by this problem, we propose salient instance segmentation, which simultaneously detects salient regions and identifies object instances inside them. Because salient object detection is not associated with a predefined set of semantic categories, it is a challenging problem closely related to generic object detection and segmentation.
We believe solutions to such generic problems are valuable in practice as it is not possible to enumerate all object categories and prepare pixel-level training data for each of them.

\section{Salient Instance Segmentation}
As shown in Fig.~\ref{fig:architecture}, our method for salient instance segmentation consists of four cascaded components, including salient region detection, salient object contour detection, salient instance generation and salient instance refinement. Specifically, we propose a deep multiscale refinement network and apply it to both salient region detection and salient object contour detection. Next, we generate a fixed number of salient object proposals on the basis of the results of salient object contour detection and apply a subset optimization method for further screening these object proposals. Finally, the results from the previous three steps are integrated in a CRF model to generate the final salient instance segmentation.

\subsection{Multiscale Refinement Network}
We formulate both salient region detection and salient object contour detection as a binary pixel labeling problem.
Fully convolutional networks have been widely used in image labeling problems and have achieved great successes in salient region detection~\cite{LiYu16,liu2016dhsnet,wang2016saliency} and object contour detection~\cite{xie2015holistically,yang2016object}. However, none of them addresses these two problems in a unified network architecture. Since salient objects could have different scales, we propose a multiscale refinement network~(MSRNet) for both salient region detection and salient object contour detection. MSRNet is composed of three refined VGG network streams with shared parameters and a learned attentional model for fusing results at different scales.

\subsubsection{Refined VGG Network}
\begin{figure}[t]
\centerline{
   \includegraphics[width=0.48\textwidth]{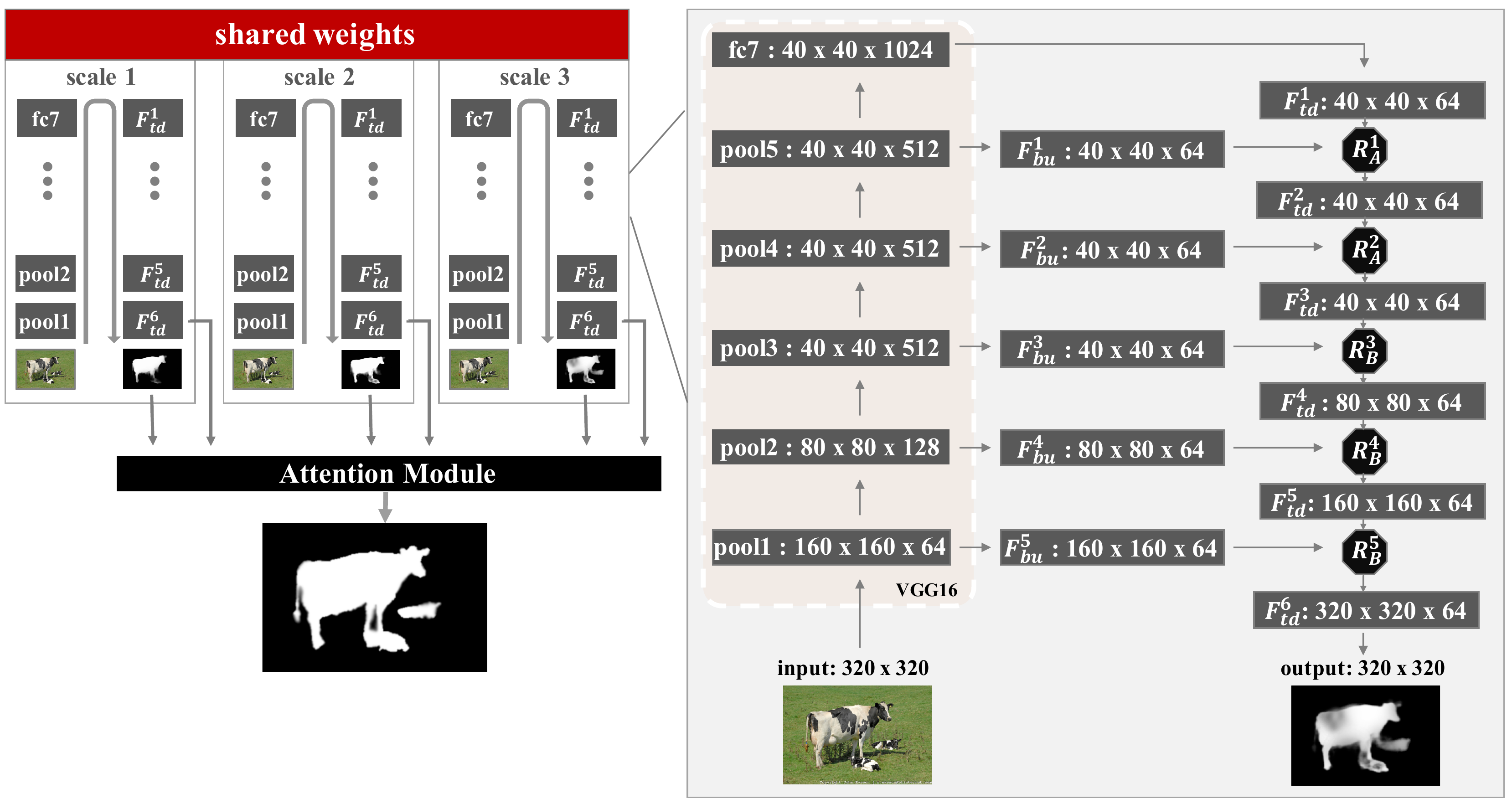}
}
   \caption{The architecture of our multiscale refinement network.\vspace{-2mm}}
\label{fig:MSRNet}
\end{figure}

Salient region detection and salient object contour detection are closely related and both of them require low-level cues as well as high-level semantic information.
Information from an input image needs to be passed from the bottom layers up in a deep network before being transformed into high-level semantic information. Meanwhile, such high-level semantic information also needs to be passed from the top layers down and further integrated with high-resolution low-level cues, such as colors and textures, to produce high-precision region and contour detection results.
Therefore, a network should consider both bottom-up and top-down information propagation and output a label map with the same resolution as the input image. We propose a refined VGG network architecture to achieve this goal. As shown in Fig.~\ref{fig:MSRNet}, the refined VGG network is essentially a VGG network augmented with a top-down refinement process.

We transform the original VGG16 into a fully convolutional network, which serves as our bottom-up backbone network. The two fully connected layers of VGG16 are first converted into convolutional layers with $1\times 1$ kernels as described in \cite{long2015fully}. We also skip subsampling in the last two pooling layers to make the bottom-up feature map denser and replace the convolutional layers after the penultimate pooling layer with atrous convolution in order to retain the original receptive field of the filters. Thus the output resolution of the transformed VGG network is $1/8$ of the original input resolution.

To augment the backbone network with a top-down refinement stream, we first attach one extra convolutional layer to each of the five max-pooling layers of VGG16. Each extra layer has $3\times 3$ kernels and 64 channels which play a role in dimension reduction. Inspired by \cite{pinheiro2016learning}, we integrate a ``refinement module'' $R$ to invert the effect of each pooling layer and double the resolution of its input feature map if necessary. As shown in Figure~\ref{fig:MSRNet}, the refinement stream consists of five stacked refinement modules, each of which corresponds to one pooling layer in the backbone network. Each refinement module $R^i$ takes as input the output feature map $F_{td}^{i}$ of the previous refinement module in the top-down pass along with the output feature map $F_{bu}^{i}$ of the aforementioned extra convolutional layer attached to the corresponding pooling layer in the bottom-up pass. It learns to merge the information from these inputs to produce a new feature map $F_{td}^{i+1}$, i.e. $F_{td}^{i+1}=R^i(F_{td}^{i}, F_{bu}^{i})$. The refinement module $R^i$ works by first concatenating $F_{td}^{i}$ and $F_{bu}^{i}$ and then feeding them to another $3\times 3$ convolutional layer with 64 channels. Finally, an up-sampling layer is optionally added to double the spatial resolution to guarantee that $F_{td}^{i}$ and $F_{bu}^{i}$ have the same spatial resolution. Specifically, an up-sampling layer is added in each refinement module corresponding to any of the first three pooling layers in the bottom-up pass. We denote a refinement operation without up-sampling as $R_A$ and that with up-sampling as $R_B$. Note that $F_{td}^{1}$ is the output feature map encoding from the last layer of the backbone network and serves as the input to the entire top-down refinement stream. The final output of the refinement stream is a probability map with the same resolution as the original input image.

\subsubsection{Multiscale Fusion with Attentional Weights}
As it has been widely confirmed that feeding multiple scales of an input image to networks with shared parameters are beneficial for accurately localizing objects of different scales in pixel labeling problems~\cite{farabet2013learning,chen2015attention,eigen2015predicting,lin2015efficient}, we replicate the refined VGG network in the previous section three times, each responsible for one of the scales.
An input image is resized to three different scales ($s\in \{1, 0.75, 0.5\}$). Each scale $s$ of the input image passes through one of the three replicated refined VGG networks, and comes out as a two-channel probability map in the resolution of scale $s$, denoted as $M_{c}^{s}$, where $c \in \{0,1\}$ denotes the two classes for saliency detection. The three probability maps are resized to the same resolution as the raw input image using bilinear interpolation.

\begin{figure}[t]
\centerline{
   \includegraphics[width=0.48\textwidth]{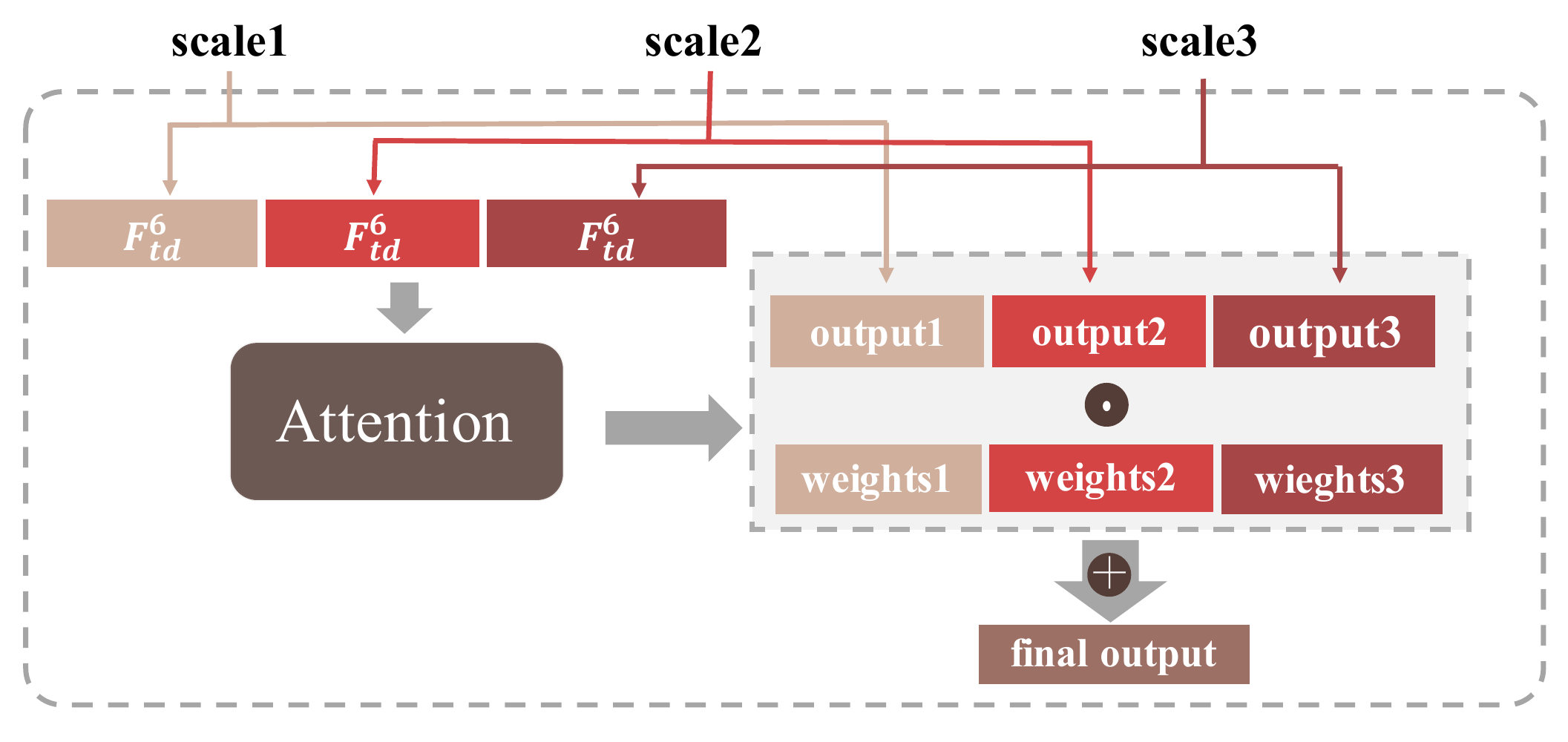}
}
   \caption{The architecture of the attention module.}
\label{fig:attention}
\end{figure}

The final output from our MSRNet is computed as a weighted sum of the three probability maps in a pixelwise manner, which means the weights for the probabilistic scores at a pixel are not fixed but spatially varying. Let $F_{c}$ be the fused probability map of class $c$ and $W^{s}$ be the weight map for scale $s$. The fused map is calculated by summing the elementwise multiplication between each probability map and its corresponding weight map:
\begin{equation}
  F_{c} = \Sigma_{s\in \{1,0.75,0.5\}}W^{s} \odot M_{c}^{s}.
\end{equation}
We call $W^{s}$ attentional weights as in \cite{gregor2015draw} because it reflects how much attention should be paid to features at different spatial locations and image scales.

These spatially varying attentional weights can be viewed as probability maps themselves and can be learned in a fully convolutional network as well. We simultaneously learn attentional weights along with saliency maps by adding an attention module to our MSRNet. As shown in Fig.~\ref{fig:attention}, the attention module takes as input the concatenation of three output feature maps of the penultimate layers in the three top-down refinement streams, and it consists of two convolutional layers for attentional weight inference. The first convolutional layer has 512 channels with $3\times 3$ kernels and the second layer has three channels with $1\times 1$ kernels. Each of the three channels in the output feature map corresponds to attentional weights for one of the three scales. Thus the attention module learns a soft weight for each spatial location and each scale. As the convolutions and elementwise multiplications in our attention module are differentiable, they allow the gradient of the loss function to be propagated through. Therefore, the attention module can be jointly trained in our MSRNet.

\subsubsection{Multiscale Refinement Network Training}
\label{sec:msrnettraining}
We train two deep models based on the same multiscale refinement network architecture to perform two subtasks, salient region detection and salient object contour detection. These subtasks have separate training sets. As the number of training images for salient contour detection is much smaller, in practice, we first train a network for salient region detection. A duplicate of this trained network is further fine-tuned for salient contour detection. The loss functions of these two subtasks have different weights for sample balance. As the number of ``contour'' and ``non-contour'' pixels are extremely imbalanced in each training batch for salient object contour detection, the penalty for misclassifying ``contour'' pixels is $10$ times the penalty for misclassifying ``non-contour'' pixels while, for salient region detection, the penalty for misclassifying ``salient'' pixels is twice the penalty for misclassifying ``non-salient'' pixels. When training MSRNet for salient region detection, we initialize the bottom-up backbone network with a VGG16 network pretrained on ImageNet and the top-down refinement stream with random values. We jointly fine-tune the three refined VGG networks in MSRNet, and their shared parameters are optimized using standard stochastic gradient descent. The learning rate for the backbone networks is set to $10^{-4}$ while that for other newly added layers is set to $10^{-3}$. To save memory and increase the mini-batch size, we fix the resolution of training images to $320\times 320$. However, as MSRNet is a fully convolutional network, it can take an image of any size as the input and produce a saliency map with the same resolution as the input during testing.

\subsection{Salient Instance Proposal}\label{sec:salientproposal}
We choose the multiscale combinatorial grouping (MCG) algorithm~\cite{arbelaez2014multiscale} to generate salient object proposals from the detected salient object contours. MCG is a unified approach for bottom-up hierarchical image segmentation and object candidate generation. We simply replace the contour detector gPb in MCG with our MSRNet based salient object contour detector. Specifically, given an input image, we first generate four salient object contour maps (three from scaled versions of the input and one from the fused map). Each of these four contour maps is used to generate a distinct hierarchical image segmentation represented as an ultrametric contour map~(UCM). These four hierarchies are aligned and combined into a single hierarchical segmentation, and a ranked list of object proposals are obtained as in~\cite{arbelaez2014multiscale}. 

To ensure a high recall rate of salient object instances, we generate 800 salient object proposals for any given image. We discard those proposals with fewer than $80\%$ salient pixels to guarantee that any remaining proposal mostly resides inside a detected salient region. Given the set of initially screened salient object proposals, we further apply a MAP-based subset optimization method proposed in \cite{zhangunconstrained} to produce a compact set of object proposals. The number of remaining object proposals in the compact set forms the final number of predicted salient object instances in the image. We call each remaining salient object proposal a detected salient instance. We can easily obtain an initial result for salient instance segmentation by labeling the pixels in each salient instance with a unique instance id.

\subsection{Refinement of Salient Instance Segmentation}
As salient object proposals and salient regions are obtained independently, there exist discrepancies between the union of all detected salient instances and the union of all detected salient regions. In this section, we propose a fully connected CRF model to refine the initial salient instance segmentation result.

Suppose the number of salient instances is $K$. We treat the background as the $K+1^{st}$ class, and cast salient instance segmentation as a multi-class labeling problem. At the end, every pixel is assigned with one of the $K+1$ labels using a CRF model. To achieve this goal, we first define a probability map with $K+1$ channels, each of which corresponds to the probability of the spatial location being assigned with one of the $K+1$ labels. If a salient pixel is covered by a single detected salient instance, the probability of the pixel having the label associated with that salient instance is 1. If a salient pixel is not covered by any detected salient instance, the probability of the pixel having any label is $\frac{1}{K}$. Note that salient object proposals may have overlaps and some object proposals may occupy non-salient pixels. If a salient pixel is covered by $k$ overlapping salient instances, the probability of the pixel having a label associated with one of the $k$ salient instances is $\frac{1}{k}$. If a background pixel is covered by $k$ overlapping salient instances, the probability of the pixel having a label associated with one of the $k$ salient instances is $\frac{1}{k+1}$, and the probability of the pixel having the background label is also $\frac{1}{k+1}$.

Given this initial salient instance probability map, we employ a fully connected CRF model~\cite{krahenbuhl2012efficient} for refinement. Specifically, pixel labels are optimized with respect to the following energy function of the CRF:
\begin{equation}
E\left( x \right) = -\sum_{i}\log P\left(x_i\right)+\sum_{i,j}\theta_{ij}\left(x_i, x_j\right),
\end{equation}
where $x$ represents a complete label assignment for all pixels and $P\left(x_i\right)$ is the probability of pixel $i$ being assigned with the label prescribed by $x$. $\theta_{ij}\left(x_i,x_j\right)$ is a pairwise potential defined as follows,
\begin{equation}
\begin{split}
\theta_{ij}=\mu\left(x_i,x_j\right)\Bigg[ \omega_1\exp\Bigg(-\frac{\left \|p_i-p_j  \right \|^2}{2\sigma_\alpha^2}-\frac{\left \|I_i-I_j \right \|^2}{2\sigma_\beta^2}\Bigg)  + \\\omega_2\exp\left(-\frac{\left \|p_i-p_j \right\|^2}{2\sigma_\gamma^2}\right)\Bigg],
\end{split}
\end{equation}
where $\mu\left(x_i,x_j\right) = 1$ if $x_i \neq x_j$, and zero otherwise. $\theta_{ij}$ involves two kernels. The first kernel depends on pixel positions ($p$) and pixel intensities ($I$), and encourages nearby pixels with similar colors to take similar salient instance labels, while the second kernel only considers spatial proximity when enforcing smoothness. The hyperparameters, $\sigma_\alpha$, $\sigma_\beta$ and $\sigma_\gamma$, control the scale of Gaussian kernels. In this paper, we apply the publicly available implementation of \cite{krahenbuhl2012efficient} to minimize the above energy. The parameters in this CRF are determined through cross validation on the validation set of our dataset introduced in the next section. The actual values of $w_1$, $w_2$, $\sigma_\alpha$, $\sigma_\beta$ and $\sigma_\gamma$ are set to $4.0$, $3.0$, $49.0$, $5.0$ and $3.0$, respectively in our experiments.

\begin{figure*}[ht]
\centerline{
   \includegraphics[width=0.95\textwidth]{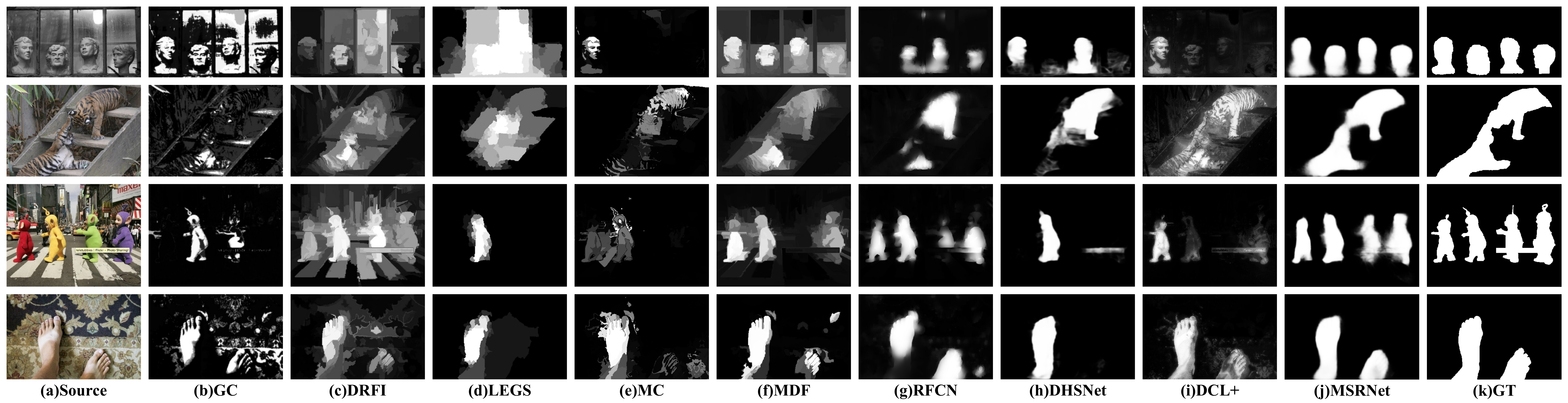}\vspace{-1mm}
}
   \caption{Visual comparison of saliency maps from state-of-the-art methods, including our MSRNet. The ground truth (GT) is shown in the last column. MSRNet consistently produces saliency maps closest to the ground truth.\vspace{-0mm}
   }
\label{fig:smaps}
\end{figure*}

\section{A New Dataset for Salient Object Instances}
As salient instance segmentation is a completely new problem, no suitable datasets exist. In order to promote the study of this problem, we have built a new dataset with pixelwise salient instance labels. We initially collected $1,388$ images. To reduce the ambiguity in salient region detection results, these images were mostly selected from existing datasets for salient region detection, including ECSSD~\cite{yan2013hierarchical}, DUT-OMRON~\cite{yang2013saliency}, HKU-IS~\cite{LiYu15}, and MSO datasets~\cite{zhangunconstrained}. Two-thirds of the chosen images contain multiple occluded salient object instances while the remaining one-third consists of images with no salient regions, a single salient object instance or multiple salient instances without occlusion. To reduce label inconsistency, we asked three human annotators to label detected salient regions with different instance IDs in all selected images using a custom designed interactive segmentation tool. We only kept the images where salient regions were divided into an identical number of salient object instances by all the three annotators. At the end, our new salient instance dataset contains 1,000 images with high-quality pixelwise salient instance labeling as well as salient object contour labeling. We randomly divide the dataset into three parts, including 500 for training, 200 for validation and 300 for testing.

\section{Experimental Results}

\subsection{Implementation}
Our proposed MSRNet has been implemented on the public DeepLab code base~\cite{chen2014semantic}, which was implemented in the Caffe framework~\cite{jia2014caffe}. A GTX Titan X GPU is used for both training and testing.
We combine the training sets of both the MSRA-B dataset (2500 images)~\cite{liu2011learning} and the HKU-IS dataset (2500 images)~\cite{LiYu15} as our training set (5000 images) for salient region detection. The validation sets in the aforementioned two datasets are also combined as our validation set~(1000 images). We augment the image dataset by horizontal flipping. During training, the mini-batch size is set to $6$ and we choose to update the loss every $5$ iterations. We set the momentum parameter to 0.9 and the weight decay to 0.0005 for both subtasks. The total number of iteration is set to $20K$. We test the softmax loss on the validation set every 500 iterations and select the model with the lowest validation loss as the best model for testing. As discussed in Section~\ref{sec:msrnettraining}, this trained model is used as the initial model for salient contour detection, and is further fine-tuned on the training set of our new dataset for salient instances and contour detection. As our new dataset only contains $500$ training images, we perform data augmentation as in \cite{xie2015holistically}. Specifically, we rotate the images to 8 different orientations and crop the largest rectangle in the rotated image. With horizontal flipping at each orientation, the training set is enlarged by 16 times. We fine-tune MSRNet on the augmented dataset for $10K$ iterations and keep the model with the lowest validation error as our final model for salient object contour detection.


It takes around 50 hours to train our multiscale refinement network for salient region detection and another 20 hours for salient object contour detection. As MSRNet is a fully convolutional network, the testing phase is very efficient. In our experiments, it takes 0.6 seconds to perform either salient region detection or salient object contour detection on a testing image with 400x300 pixels. It takes 20 seconds to generate a salient instance segmentation with MCG being the bottleneck which needs 18 seconds to generate salient object proposals for a single image.

\begin{figure*}[ht]
    \centerline{
    \includegraphics[width = 0.32\textwidth]{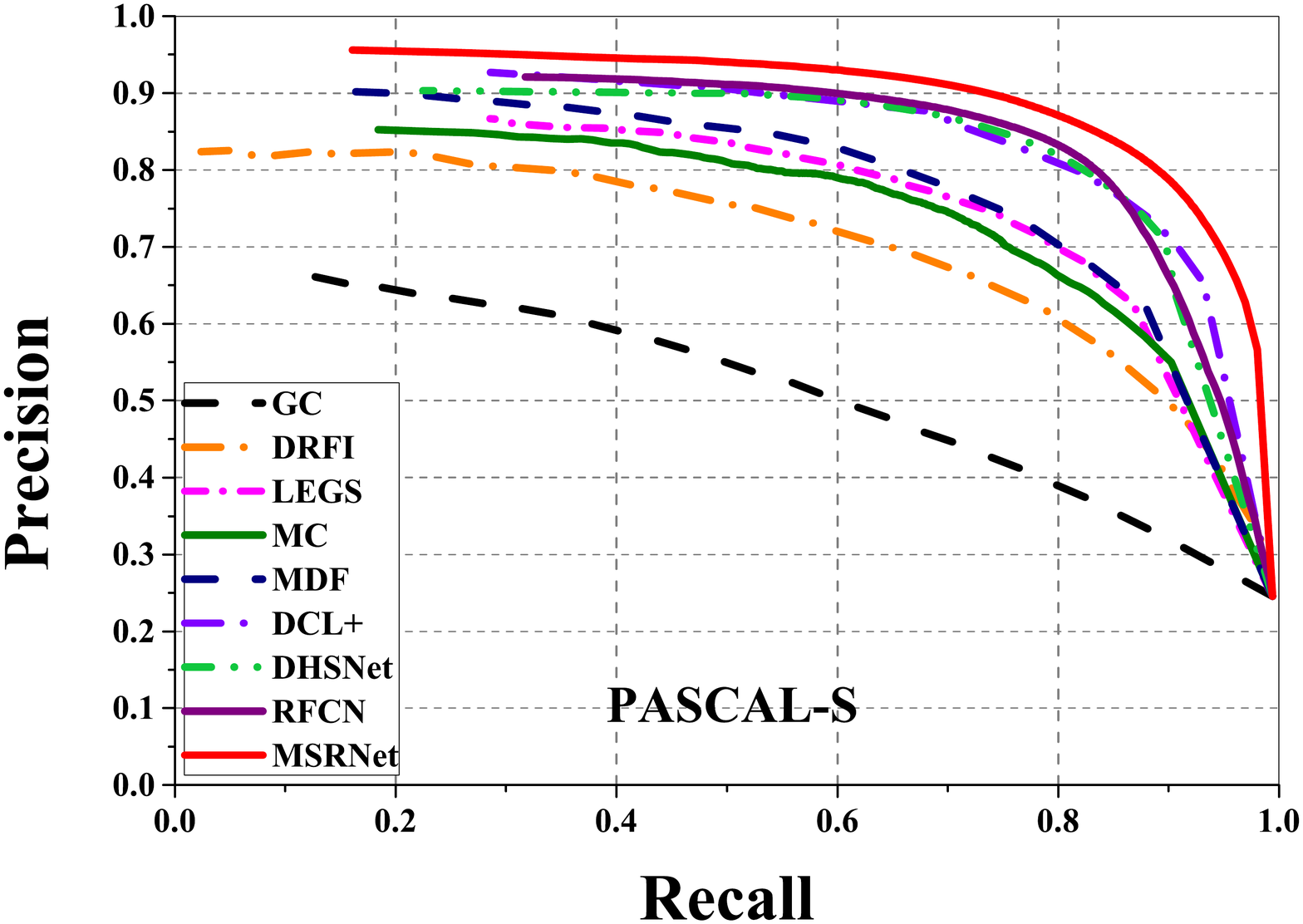}\hfill
    \includegraphics[width = 0.32\textwidth]{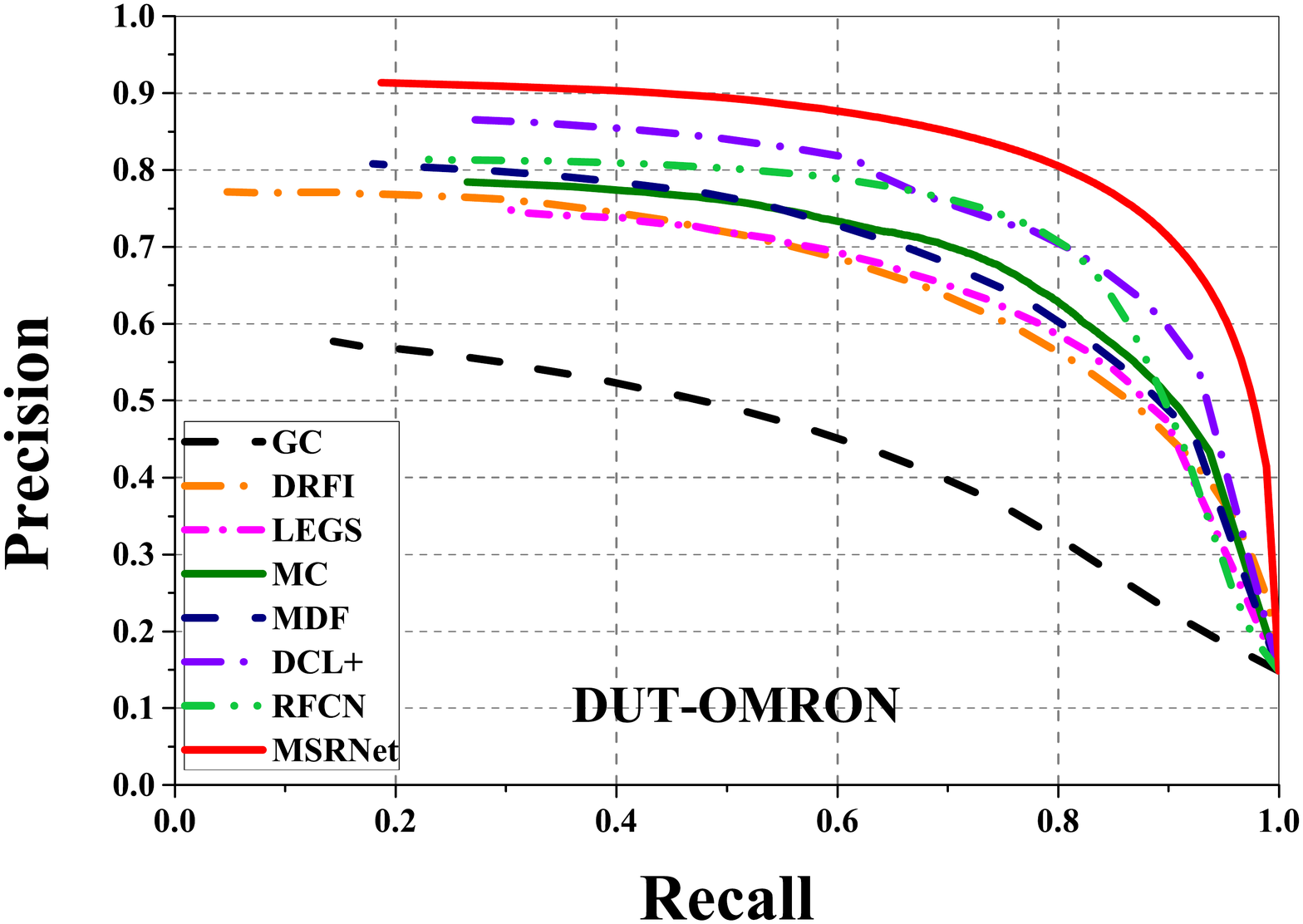}\hfill
    \includegraphics[width = 0.32\textwidth]{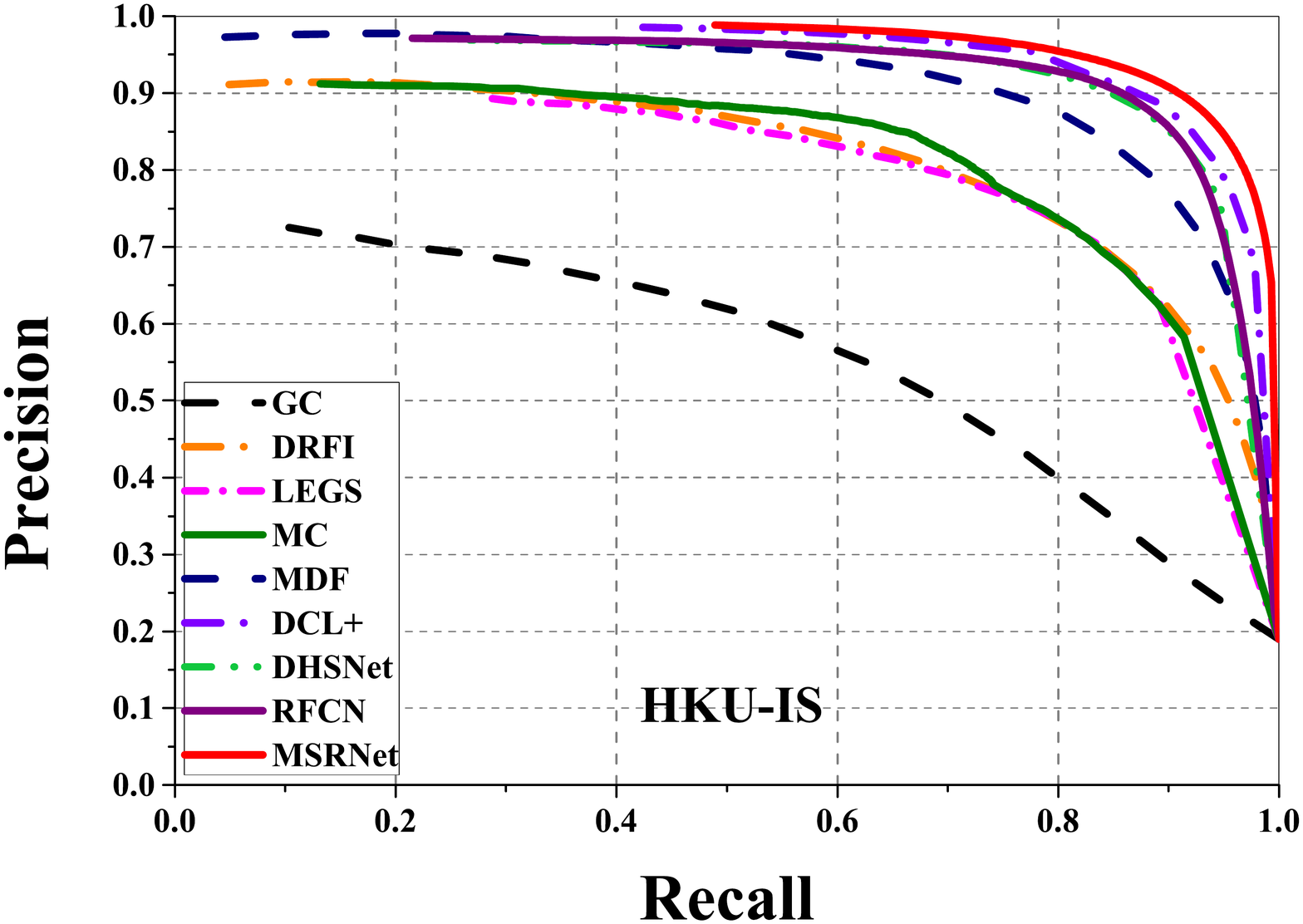}
  }
  \caption{Comparison of precision-recall curves among 9 salient region detection methods on 3 datasets. Our MSRNet consistently outperforms other methods across all the testing datasets. Note that DHSNet~\cite{liu2016dhsnet} includes the testing set of DUT-OMRON in its training data, therefore DHSNet is not included in the comparison on this dataset.\vspace{-2mm}
  }
  \label{fig:comps_pr}
\end{figure*}

\subsection{Evaluation on Salient Region Detection}
To evaluate the performance of our MSRNet on salient region detection, we conduct testing on six benchmark datasets: MSRA-B~\cite{liu2011learning}, PASCAL-S~\cite{li2014secrets}, DUT-OMRON\cite{yang2013saliency}, HKU-IS~\cite{LiYu15}, ECSSD~\cite{yan2013hierarchical} and SOD~\cite{martin2001database}.
As we train our network on the combined training sets of MSRA-B and HKU-IS,
we evaluate our trained model on the testing sets of these two datasets and on the combined training and testing sets of other datasets.

We adopt precision-recall curves~(PR), maximum F-measure and mean absolute error~(MAE) as our performance measures.
The F-measure is defined as
  $F_{\beta} = \frac{(1+\beta^2)\cdot Precision \cdot Recall}{\beta^2\cdot Precision + Recall},$
where $\beta^2$ is set to 0.3. We report the maximum F-measure computed from all precision-recall pairs.
MAE is defined as the average pixelwise absolute difference between the binary ground truth and the saliency map~\cite{perazzi2012saliency}.
It is a more meaningful measure in evaluating the applicability of a saliency model in salient instance segmentation.
In the supplemental materials, we also report the average precision, recall and F-measure using an adaptive threshold which is set to twice the mean saliency value of each saliency map as suggested in~\cite{achanta2009frequency}.

\begin{table*}[]
\centering
\resizebox{0.88\textwidth}{!}
{
\begin{tabular}{|c|c|c|c|c|c|c|c|c|c|c|}
\hline
Dataset                    & Metric & GC    & DRFI  & LEGS                         & MC                                    & MDF                          & RFCN                                  & DHSNet                                & DCL+                                  & MSRNet                                \\ \hline
                            & maxF   & 0.719 & 0.845 & {\color[HTML]{333333} 0.870} & {\color[HTML]{32CB00} \textbf{0.894}} & 0.885                        & ---                                   & ---                                   & {\color[HTML]{3531FF} \textbf{0.916}} & {\color[HTML]{FE0000} \textbf{0.930}} \\ \cline{2-11}
\multirow{-2}{*}{MSRA-B}    & MAE    & 0.159 & 0.112 & {\color[HTML]{333333} 0.081} & {\color[HTML]{32CB00} \textbf{0.054}} & 0.066                        & ---                                   & ---                                   & {\color[HTML]{3531FF} \textbf{0.047}} & {\color[HTML]{FE0000} \textbf{0.042}} \\ \hline
                            & maxF   & 0.539 & 0.690 & {\color[HTML]{333333} 0.752} & 0.740                                 & 0.764                        & {\color[HTML]{3531FF} \textbf{0.832}} & {\color[HTML]{32CB00} \textbf{0.824}} & {\color[HTML]{333333} 0.822}          & {\color[HTML]{FE0000} \textbf{0.852}} \\ \cline{2-11}
\multirow{-2}{*}{PASCAL-S}  & MAE    & 0.266 & 0.210 & {\color[HTML]{333333} 0.157} & 0.145                                 & 0.145                        & 0.118                                 & {\color[HTML]{3531FF} \textbf{0.094}} & {\color[HTML]{32CB00} \textbf{0.108}} & {\color[HTML]{FE0000} \textbf{0.081}} \\ \hline
                            & maxF   & 0.495 & 0.664 & {\color[HTML]{333333} 0.669} & {\color[HTML]{333333} 0.703}          & 0.694                        & {\color[HTML]{32CB00} \textbf{0.747}} & ---                                    & {\color[HTML]{3531FF} \textbf{0.757}} & {\color[HTML]{FE0000} \textbf{0.785}} \\ \cline{2-11}
\multirow{-2}{*}{DUT-OMRON} & MAE    & 0.218 & 0.150 & {\color[HTML]{333333} 0.133} & {\color[HTML]{32CB00} \textbf{0.088}} & {\color[HTML]{333333} 0.092} & 0.095                                 & ---                                    & {\color[HTML]{3531FF} \textbf{0.080}} & {\color[HTML]{FE0000} \textbf{0.069}} \\ \hline
                            & maxF   & 0.588 & 0.776 & {\color[HTML]{333333} 0.770} & 0.798                                 & 0.861                        & {\color[HTML]{32CB00} \textbf{0.896}} & 0.892                                 & {\color[HTML]{3531FF} \textbf{0.904}} & {\color[HTML]{FE0000} \textbf{0.916}} \\ \cline{2-11}
\multirow{-2}{*}{HKU-IS}    & MAE    & 0.211 & 0.167 & {\color[HTML]{333333} 0.118} & 0.102                                 & {\color[HTML]{333333} 0.076} & 0.073                                 & {\color[HTML]{32CB00} \textbf{0.052}} & {\color[HTML]{3531FF} \textbf{0.049}} & {\color[HTML]{FE0000} \textbf{0.039}} \\ \hline
                            & maxF   & 0.597 & 0.782 & 0.827                        & 0.837                                 & 0.847                        & 0.899                                 & {\color[HTML]{3531FF} \textbf{0.907}} & {\color[HTML]{32CB00} \textbf{0.901}} & {\color[HTML]{FE0000} \textbf{0.913}} \\ \cline{2-11}
\multirow{-2}{*}{ECSSD}     & MAE    & 0.233 & 0.170 & 0.118                        & 0.100                                 & 0.106                        & 0.091                                 & {\color[HTML]{3531FF} \textbf{0.059}} & {\color[HTML]{32CB00} \textbf{0.068}} & {\color[HTML]{FE0000} \textbf{0.054}} \\ \hline
                            & maxF   & 0.526 & 0.699 & {\color[HTML]{333333} 0.732} & 0.727                                 & 0.785                        & 0.805                                 & {\color[HTML]{32CB00} \textbf{0.823}} & {\color[HTML]{3531FF} \textbf{0.832}} & {\color[HTML]{FE0000} \textbf{0.847}} \\ \cline{2-11}
\multirow{-2}{*}{SOD}       & MAE    & 0.284 & 0.223 & {\color[HTML]{333333} 0.195} & 0.179                                 & {\color[HTML]{333333} 0.155} & 0.161                                 & {\color[HTML]{32CB00} \textbf{0.127}} & {\color[HTML]{3531FF} \textbf{0.126}} & {\color[HTML]{FE0000} \textbf{0.112}} \\ \hline
\end{tabular}
}
\caption{Comparison of quantitative results including maximum F-measure (larger is better) and MAE (smaller is better). The best three results on each dataset are shown in \color[HTML]{FE0000}\textbf{red}, \color[HTML]{3166FF}\textbf{blue}\color{black}, and \color[HTML]{32CB00}\textbf{green} \color{black}, respectively. Note that the training set of DHSNet~\cite{liu2016dhsnet} includes the testing set of MSRA-B and Dut-OMRON, and the entire MSRA-B dataset is used as part of the training set of RFCN~\cite{wang2016saliency}. Corresponding test results are excluded here.\vspace{-3mm}
}
\label{tab:comp_quantity}
\end{table*}


\subsubsection{Comparison with the State of the Art}
We compare the proposed MSRNet with other $8$ state-of-the-art salient region detection methods, including GC~\cite{cheng2015global}, DRFI~\cite{jiang2013salient}, LEGS~\cite{wang2015deep}, MC~\cite{zhao2015saliency}, MDF~\cite{LiYu15}, DCL$^+$~\cite{LiYu16}, DHSNet~\cite{liu2016dhsnet} and RFCN~\cite{wang2016saliency}. The last six are the latest deep learning based methods. We use the original implementations provided by the authors in this comparison.

A visual comparison is given in Fig.~\ref{fig:smaps}. As we can see, our proposed MSRNet can not only accurately detect salient objects at different scales but also generate more precise saliency maps in various challenging cases.
As a part of quantitative evaluation, we show a comparison of PR curves in Fig.~\ref{fig:comps_pr}.
Refer to the supplemental materials for the performance comparison on the MSRA-B, ECSSD and SOD datasets. Furthermore, a quantitative comparison of maximum F-measure and MAE is given in Table~\ref{tab:comp_quantity}. As shown in Fig.~\ref{fig:comps_pr} 
and Table~\ref{tab:comp_quantity}, our proposed MSRNet consistently outperforms existing methods across all the datasets with a considerable margin. Specifically, MSRNet improves the maximum F-measure achieved by the best-performing existing algorithm by 1.53\%, 1.33\%, 3.70\%, 1.33\%, 2.4\% and 1.8\% respectively
on MSRA-B, HKU-IS, DUT-OMRON, ECSSD, PASCAL-S and SOD. And at the same time, MSRNet lowers the previoiusly best MAE by 10.6\%, 20.4\%, 13.8\%, 8.5\%, 13.8\% and 11.1\% respectively on MSRA-B, HKU-IS, DUT-OMRON, ECSSD, PASCAL-S and SOD. It is worth noting that MSRNet outperforms all the other six deep learning based saliency detection methods without resorting to any post-processing techniques such as CRF.

\subsubsection{Effectiveness of Multiscale Refinement Network}
\begin{figure}[h]
    \centerline{
	\includegraphics[width = 0.222\textwidth]
    {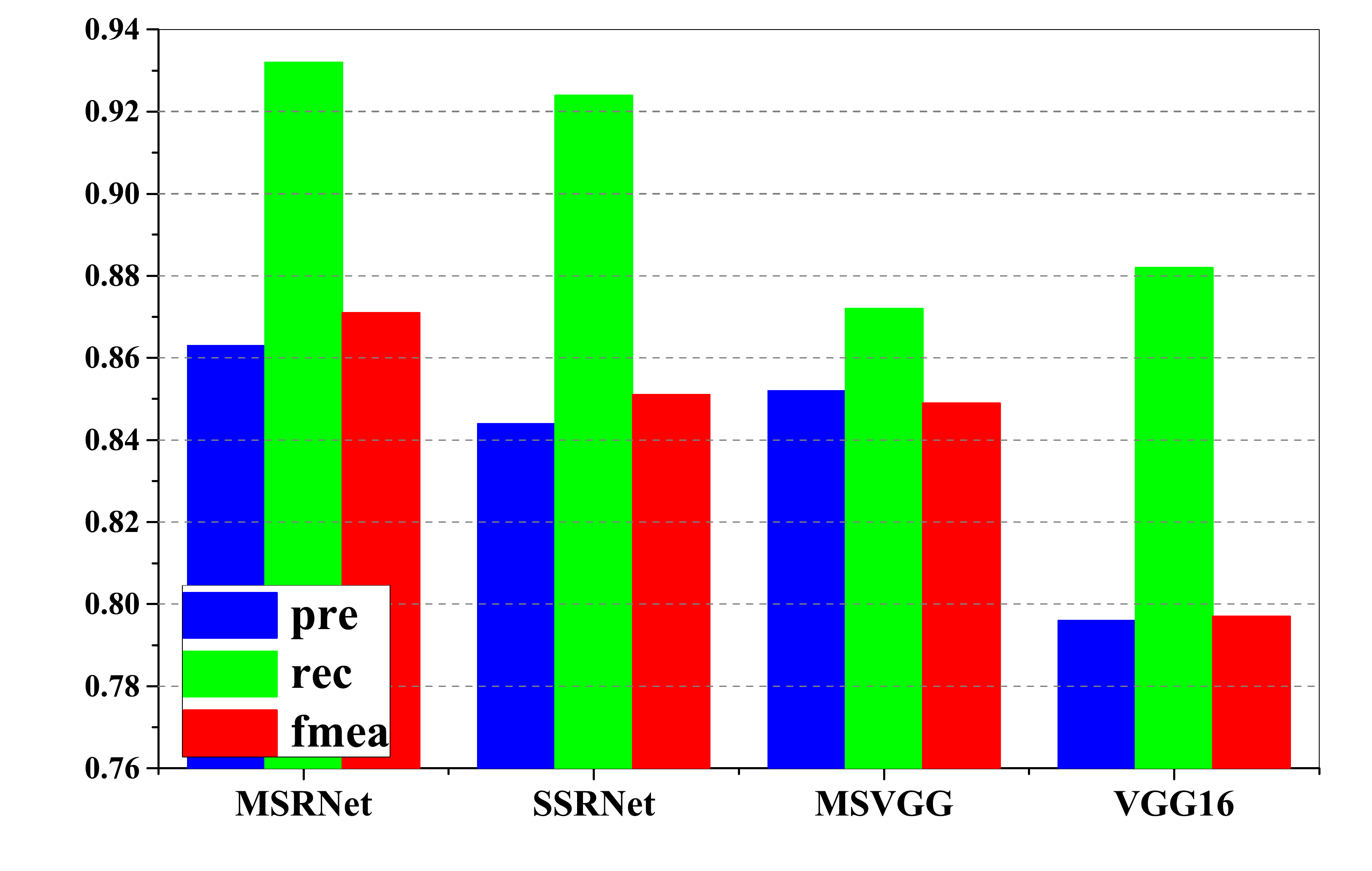}\hfill
    \includegraphics[width = 0.222\textwidth]
    {graphs/interp_prf-eps-converted-to.pdf}\hfill
  }\vspace{-0mm}
    \caption{Componentwise efficacy of the proposed multiscale refinement network.\vspace{-3mm}}
\label{fig:ablation_study}
\end{figure}

Our proposed MSRNet consists of three refined VGG streams and a learned attentional model for fusing results at different scales. To demonstrate the effectiveness and necessity of each component, we have trained three additional models for comparison. These three models are respectively a single backbone network (VGG16), a single-scale refinement network (SSRNet) and a multiscale VGG network with the same attentional module but without refinement (MSVGG). These three additional models are trained using the same setting as MSRNet training. Quantitative results from the four methods are obtained on the testing part of HKU-IS dataset. As shown in Fig.~\ref{fig:ablation_study}, MSRNet consistently achieves the best performance in terms of the PR curve as well as average precision, recall and F-measure. Both SSRNet and MSVGG perform much better than VGG16, which respectively demonstrates the effectiveness of the refinement module and attention based multiscale fusion in MSRNet. Moreover, these two components are complementary to each other, which makes MSRNet not only capable of detecting more precise salient regions (with higher resolution) but also discovering salient objects at multiple scales.

\subsection{Evaluation on Salient Instance Segmentation}
To evaluate the effectiveness of our proposed framework for salient instance segmentation as well as to promote further research on this new problem, we adopt two types of performance measures and demonstrate the results from our framework according to these measures.

We use the same performance measures used for traditional contour detection~\cite{arbelaez2011contour,xie2015holistically} to evaluate the performance of salient object contour detection, and adopt three standard measures: fixed contour threshold (ODS), per-image best threshold (OIS), and average precision (AP). Refer to~\cite{arbelaez2011contour} for detailed definitions.
We define performance measures for salient instance segmentation by drawing inspirations from the evaluation of instance-aware semantic segmentation. Specifically, we adopt mean Average Precision, referred to as $mAP^{r}$~\cite{hariharan2014simultaneous}. In this paper, we report $mAP^{r}$ using IoU thresholds at 0.5 and 0.7, denoted as $mAP^{r}@0.5$ and $mAP^{r}@0.7$ respectively.

\begin{figure}[ht]
\begin{center}
   \includegraphics[width=0.43\textwidth]{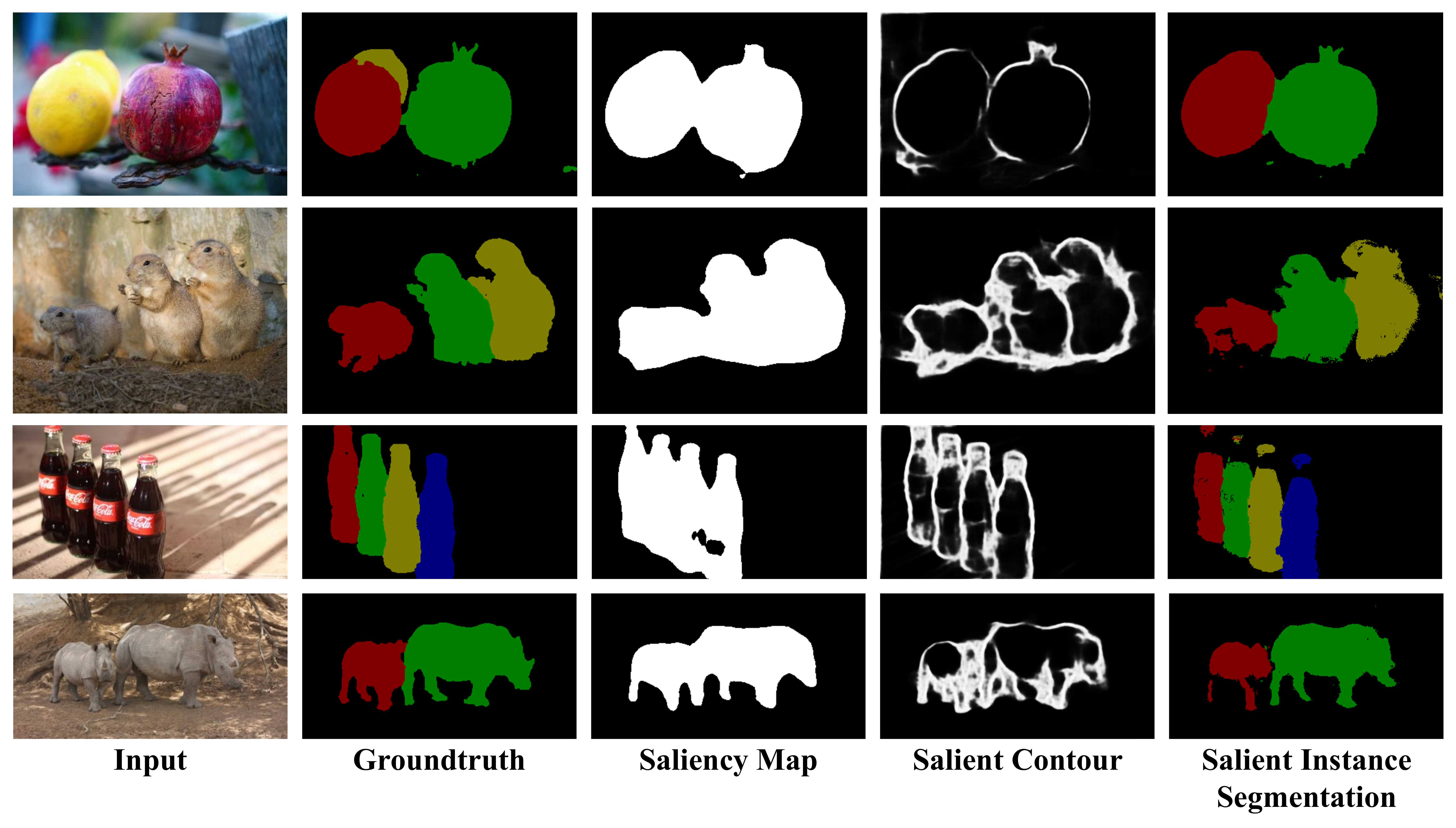}\vspace{-5mm}
\end{center}
   \caption{Examples of salient instance segmentation results by our MSRNet based framework. 
   }
\label{fig:salient_instance}
\vspace{-2mm}
\end{figure}

Benchmark results from our proposed method in both salient object contour detection and salient instance segmentation are given in Table~\ref{tab:benchmark}. Fig.~\ref{fig:salient_instance} demonstrates examples from our results on our testing set. Our method can handle challenging cases where multiple salient object instances are spatially connected to each other.


\begin{table}[h]
\centering
\caption{Quantitative benchmark results of salient object contour detection and salient instance segmentation on our new dataset.}
\label{tab:benchmark}
\resizebox{0.45\textwidth}{!}
{
\begin{tabular}{ccc||cc}
\hline
\multicolumn{3}{c}{\textbf{Salient Contour Detection}} & \multicolumn{2}{l}{\textbf{Salient Instance Segmentation}} \\ \hline
ODS              & OIS              & AP               & $MP^r$@0.5(\%)                  & $MP^{r}$@0.7(\%)                 \\ \hline
0.719            & 0.757            & 0.765            & 65.32                          & 52.18                         \\ \hline
\end{tabular}

}

\end{table}

\section{Conclusions}
In this paper, we have introduced salient instance segmentation, a new problem related to salient object detection, and also presented a framework for solving this problem. The most important component of our framework is a multiscale saliency refinement network, which generates high-quality salient region masks and salient object contours. To promote further research and evaluation of salient instance segmentation, we have also constructed a new database with pixelwise salient instance annotations. Experimental results demonstrate that our proposed method is capable of achieving state-of-the-art performance on all public datasets for salient region detection as well as on our new dataset for salient instance segmentation.

\section*{Acknowledgment}
This work was supported by Hong Kong Innovation and Technology Fund (ITP/055/14LP), State Key Development Program under Grant 2016YFB1001004, the National Natural Science Foundation of China under Grant 61622214 and Special Program of the NSFC-Guangdong Joint Fund for Applied Research on Super Computation (the second phase).

{\small
\bibliographystyle{ieee}
\bibliography{instancesaliency}
}

\end{document}